\documentclass{article}

\usepackage{PRIMEarxiv}

\usepackage[utf8]{inputenc} 
\usepackage[T1]{fontenc}    
\usepackage{hyperref}       
\usepackage{url}            
\usepackage{booktabs}       
\usepackage{amsfonts}       
\usepackage{nicefrac}       
\usepackage{microtype}      
\usepackage{lipsum}
\usepackage{fancyhdr}       
\usepackage{graphicx}       
\graphicspath{{media/}}     

\usepackage{mathtools} 
\usepackage{amsmath,amssymb}
\usepackage{algorithmic}
\usepackage{graphicx}
\usepackage{textcomp}
\usepackage{xcolor}
\usepackage[utf8]{inputenc}
\usepackage{hyperref}
\usepackage{float}
\usepackage{bbold}
\usepackage{algorithm}
\usepackage{algorithmic}
\usepackage{comment}
\usepackage{enumitem}
\usepackage{caption}
\usepackage{subcaption}
\usepackage[sectionbib]{natbib}
\usepackage{cite}
\def \si {\sigma}

\def \be*{\begin{eqnarray*}}
\def \e*{\end{eqnarray*}}
\def \beg{\begin{eqnarray}}
\def \en{\end{eqnarray}}

\def \bit{\begin{itemize}}
\def \eit{\end{itemize}}

\def \w {\widehat}

\title{Data Augmentation with Variational Autoencoder for Imbalanced Dataset}

\author{
  Samuel Stocksieker \\
  Aix Marseille Université\\
  I2M, CNRS, Centrale Marseille\\ 
  Marseille, 
  France \\
   \And
  Denys Pommeret \\
  Aix Marseille Université\\
  I2M, CNRS, Centrale Marseille\\ 
  Marseille, 
  France \\
   \And
  Arthur Charpentier \\
  Université du Québec à Montréal\\
  Département de Mathématique\\
  Montréal, 
  Canada
}

\begin{document}
\maketitle

\begin{abstract}
Learning from an imbalanced distribution presents a major challenge in predictive modeling, as it generally leads to a reduction in the performance of standard algorithms. Various approaches exist to address this issue, but many 
of them concern classification problems, with a limited focus on regression. In this paper, we introduce a novel method aimed at enhancing learning on tabular data in the Imbalanced Regression (IR) framework, which remains a significant problem. We propose to use variational autoencoders (VAE) which are known as a powerful tool for synthetic data generation, offering an interesting approach to modeling and capturing latent representations of complex distributions. However, VAEs can be inefficient when dealing with IR. Therefore, we develop a novel approach for generating data, combining  VAE with a smoothed bootstrap, specifically designed to address the challenges of IR. 
We numerically investigate the scope of this method by comparing it against its competitors on simulations and datasets known for IR. 
\keywords{}
\end{abstract}

\section{Introduction}

Regression techniques are widely 
used 
in various fields such as finance, economics, medicine, and engineering, to model relationships between features and a continuous target variable. However, datasets in real-world applications often exhibit a significant imbalance in the distribution of the target variable, presenting specific challenges for traditional modeling methods (\citep{krawczyk2016learning}, \citep{fernandez2018learning}). 
Indeed, standard approaches typically treat all values 
as equally important and are trained to minimize the average error. Since rare and/or extreme values are few in the training set, they have little influence on error minimization during the learning process. 
But such values are often highly relevant. \\
This issue has been extensively studied in the context of classification \citep{haixiang2017learning}, \citep{fernandez2018smote}. However, Very few works have addressed the problem of Imbalance Regression (IR)  although many important real-world applications  \citep{krawczyk2016learning}. Indeed, regression poses additional challenges compared to classification: i) rare/minority values are not identifiable because the distribution is continuous; ii) the degree of imbalance is not easily identifiable; iii) the extent of rebalancing is also difficult to measure; and iv) generating synthetic data also requires generating a new value for the target variable.

For tabular data, preprocessing solutions are often preferred \citep{branco2016survey, he2013imbalanced}  for their universality: they allow the use of any standard regression technique. Preprocessing methods rely notably on resampling through synthetic data generation. Several methods have been proposed to generate synthetic observations from the initial data space, such as the well-known SMOTE algorithm \citep{Chawla2002Smote}. It is also possible to use a method that embeds observations into a latent space to generate data from it, as Variational Autoencoders (VAE). VAEs is a data generation method that utilizes a neural network to encode observations into a continuous latent space, where new data can be probabilistically generated. This approach has proven to be relatively effective in generating unstructured data, particularly due to its ability to capture nonlinear relationships. 

Our goal here is to generate artificial rare values while taking into account the observed correlations. Indeed, if we use a synthetic data generator, regression requires generating a new value for the target variable, unlike in the case of classification.
We propose to use a VAE approach to construct a new latent representation of the data, enabling a more relevant generation than directly generating from their initial space. The use of a VAE is also justified because, as we will present below, we aim to retain a representation of the observations in the latent space for specific guided generation in IR. However, it is well known that neural networks, including VAEs, are not efficient on tabular data \citep{shwartz2022tabular}. Therefore, we favor a "preprocessing" approach over "in-processing" to enable the application of any learning algorithm subsequently. 
In addition, a "preprocessing" solution enables the use of a "white box" model which is transparency and explainability. This is recommended in many fields such as insurance and finance,  but also in the social sciences or economics.  

In this paper, we propose a simple and effective modification of VAEs to adapt their training and generation to the problem of IR on structured data. 
Our main contributions can be summarized as follows: 
\bit
\item[i)] Providing a new loss function for training VAEs in the case of IR  
\item[ii)] Proposal of a new weighting scheme to control the sampling of rare values in the case where the target variable is continuous (regression) 
\item[iii)] Introduction of a new synthetic data generator by combining the learning power of VAEs and a non-parametric generator: the smoothed bootstrap. This technique has the advantage of generating data from a "seed" observation with the help of its neighborhood.
\eit
The paper is organized as follows. 
Section \ref{RelatedWorks} presents various state-of-the-art studies along with their shortcomings and differences compared to our proposal.
In Section \ref{proposition}, we present our DAVID approach combining a VAE adapted to IR  with an alternative method of generating data from a VAE. 
Numerical results are presented in Section \ref{illustration} with an illustration and in Section \ref{Experiments} with real applications. 
Finally, we discuss the proposed method  in Section \ref{Discussion}.

\section{Related works} \label{RelatedWorks}

We will distinguish two cases: 1) The imbalanced regression, which was initially addressed in the context of modeling the distributions of the target variable in structured tabular data; 2) The deep imbalanced regression, which has been introduce more recently,  extending the imbalance regression concept to unstructured data (such as images or NLP), where deep learning approaches are primarily used and quite effective.

\subsection{Imbalanced Regression}
The initial works on IR were established based on a utility function that allowed binarizing the issue. More precisely, \citep{ribeiro2011utility} first proposed using a relevance function to assign a value to each value of the target variable $Y$. Then, the majority and minority values are identified using a user-defined threshold. This initial approach allows adapting existing solutions within the framework of imbalanced classification, which are much more numerous. This is the case with the adaptation of the famous SMOTE algorithm to regression \citep{torgo2013smote}. This binary partitioning method has enabled the conversion of other synthetic data generation methods such as Gaussian Noise (\citep{branco2016ubl}, \citep{branco2017smogn}, \citep{song2022distsmogn}) or SMOTE extensions (\citep{moniz2018smoteboost}, \citep{camacho2022geometric}). Other methods are proposed on the same approach (e.g \citep{branco2019pre}, \citep{wu2022imbalancedlearningregression} or \citep{branco2018rebagg}). This primary approach has allowed for the first solutions to be proposed for IR. 
Moreover, generating data in the initial data space can present challenges such as identifying and generating nonlinear relationships or handling qualitative features. 
Recently, a new approach has also been proposed by \citep{stocksieker2023data} which focuses more on the features imbalance rather than the target variable.

\subsection{Deep Imbalanced Regression}
More recently, the issue of IR has been extended to the context of unstructured data  (for example, estimating the age of individuals from images) what is known as the \textit{Deep Imbalanced Regression}. Given that deep learning approaches are very effective on this type of data, it is natural that new approaches have emerged. \citep{yang2021} proposed the use of kernel density estimations to smooth the distributions of variables,  improving learning with a neural network when the target variable is continuous and imbalanced. The authors adapted the regression variant Focal-R proposed for classification tasks (\citep{lin2017focal}) for regression context by replacing the scaling factor by a continuous function.
\citep{ding2022deep} built upon this idea of smoothing via kernel density estimation and adds techniques of "CORrelation ALignment" and "class-balanced re-weighting" to improve the results. \citep{gong2022ranksim} proposed a method to enhance the performance of deep regression models in the case of imbalanced regression by incorporating a ranking similarity regularization into the loss function, thereby balancing the importance of different classes and improving overall model accuracy. 
\citep{ren2022balanced} demonstrated that using the standard MSE for imbalanced regression is inefficient and introduced a balanced MSE to improve learning with neural net techniques trained on images. 
\citep{keramati2023conr} proposed a \textit{contrastive regularizer} technique, from Contrastive Learning. \citep{sen2023dealing} handled the issue using logarithmic transformation and artificial neural network.  
Finally, as proposed for classification (\citep{ai2023generative}, \citep{zhang2018over}, \citep{utyamishev2019progressive}), the VAE approach has been proposed as a solution to the problem of imbalanced regression with \citep{wang2024variational}. The authors adapted the VAE for dealing with the deep imbalanced regression. Unlike the original VAE, which performs inference for each observation (assuming i.i.d latent representation), the authors suggested using the similarity between observations for the inference part of the VAE. 
However, as the previous works in Deep Imbalanced Regression, they used an arbitrary discretization of the continuous target variable (with equal-interval or equal-size). This technique introduces parameterization complexity with a new hyperparameter (number of bins), which significantly affects the modeling process. Furthermore, discretizing the target variable's support is not recommended as it may lead to information loss. We believe it is preferable to preserve the continuous nature of the distribution to distinguish the values of the variable of interest. Finally, the loss function used in their work is not suitable for imbalanced regression, which does not guarantee that the inference for these points is relevant, and aggregating the parameters to improve it may not solve the problem.  \\
Moreoever, as demonstrated by \citep{ren2022balanced}, the standard MSE is not relevant for imbalanced regression and like all standard algorithms, the VAE focuses on an average error and thus neglects the reconstruction (and therefore generation) of rare values. However, beyond their generative capacity, VAEs offer a latent space with interesting and exploitable properties: each point has a  latent representation in a regular space; that is, with continuity and completeness. Finally, neural networks have the advantage of being able to capture nonlinear correlations, which is crucial for generating synthetic data.

\section{DAVID: A Novel Synthetic Data Generator} \label{proposition}

Let $\boldsymbol{x}=(x_{ij})_{i=1,\cdot,n ; j=1,\cdot,p} \in \mathcal{X} \subset \mathbb{R}^{n \times p}$ be a dataset composed of $n$ observations and $p$ variables where $x_{ij}$ is the variable $j$ for the observation $i$. Let $y=(y_{i})_{i=1,\cdot,n} \in \mathcal{Y} \subset \mathbb{R^n}$ the associated target variable, which is imbalanced and continuous.

\subsection{$\beta$-VAE for Regression}
Autoencoders enable the construction of a latent representation space for data, capturing non-linear relationships between variables. By introducing a stochastic component into the latent space, Variational Autoencoders (VAE) train latent random variables representing the data and naturally generates new synthetic data very close to the initial data. VAEs are trained to reconstruct inputs from latent variables. However, in a regression framework, the target variable $Y$ should not be treated as a feature $X$. Therefore, we propose to construct a mixed VAE: the VAE takes both the input features $X$ and the target variable $Y$ and aims to reconstruct them, but with a weighting specific to $Y$.


The global loss function is therefore classically defined as follows:
\begin{align*}
\mathcal{L}(\theta, \phi, \boldsymbol{x}, y)= \beta_x \mathbb{E}_q[log \, p_\theta(\boldsymbol{x}|\boldsymbol{z})] - \beta_{KL} D_{KL}(q_\theta(z|\boldsymbol{x})||p_\theta(\boldsymbol{z})) + \beta_y \mathbb{E}_q [log \, p_\theta (y|\boldsymbol{z})]
\end{align*}
where \{$\beta_x$, $\beta_y$, $\beta_{KL}$\} are the weights, $\mathbb{E}_q [log \, p_\theta (\boldsymbol{x}|\boldsymbol{z})]$ (resp.  $\mathbb{E}_q [log \, p_\theta (y|\boldsymbol{z})]$) represents the  loss function reconstruction for $\boldsymbol{x}$ (resp.  $y$) and $D_{KL}(q(z|x), p(z|x))$ represents the Kullback-Leibler Divergence for regularization ($q_\theta(z|\boldsymbol{x})$  represents the distribution of latent variables $z$ given the input data $x$ and $p_\theta(\boldsymbol{z})$ represents the prior distribution, often chosen as Gaussian). The advantage of using a VAE here is to ensure the regularity (continuity of the prior) of the latent space, through penalization, to enable coherent data generation. 


\subsection{A Balanced loss function for Imbalanced Regression}
It is demonstrated that the Mean Squared Error (MSE) loss function is not effective in addressing imbalanced regression \citep{ren2022balanced}. Indeed, it mechanically favors frequent values, as they significantly reduce the loss function, neglecting rare values as a consequence. Here, we propose the use of a balanced loss function taking into account the frequency of the target variable $Y$.  Indeed, the goal is to train a model capable of generating new synthetic data for the rare values of $Y$. For classification tasks, rare values are directly identifiable as they belong to one or more minority classes. However, in regression, where the support of the target variable is continuous, defining rare values already poses an initial challenge.
Here, we propose to weigh the observations during the training phase by the inverse of the empirical density of $Y$. In other words, the rarer a value, the higher its weight, and vice versa. Finally, we suggest mitigating the weights using a parameter $\alpha$. The higher this parameter, the greater the difference between the weights will be.
The global loss function thus becomes:
\begin{equation}\label{Bal_loss}
    \mathcal{L}(\theta, \phi, \boldsymbol{x}, y)= \beta_x \mathbb{E}_q[log \, p_\theta(\boldsymbol{x}|\boldsymbol{z})] - \beta_{KL} D_{KL}(q_\theta(z|\boldsymbol{x})||p_\theta(\boldsymbol{z})) \\ +  \frac{\beta_y}{\w f(Y)^\alpha} \mathbb{E}_q [log \, p_\theta (y|\boldsymbol{z})]
\end{equation}

Since the density of $Y$ is unknown, we propose to estimate it using a kernel estimator (with a smoothing parameter ensuring its convergence e.g Silverman's bandwidth matrix, \citet{silverman1986density}, or that of 
\citet{scott2015multivariate}). We can find a comparable approach in \citep{steininger2021density}, but the authors employed the opposite rather than the inverse. Another approach, called "inverse re-weighting", is quite similar in \citep{yang2021} where the authors proposed to re-weight the loss function by multiplying it by the inverse of the Label Distribution Smoothing estimated density for each target bin i.e. by partitioning the domain of $Y$. However, partitioning the support of Y poses a risk of information loss.

\subsection{A Smoothed Bootstrap for Data Generation}
\label{smb_generation}

As previously stated, the goal and benefit of VAEs reside in their capacity to produce data closely resembling that on which the model was trained. Typically, a classical VAE (i.e. using Gaussian distributions) aims to calibrate the parameters $\mu_i$ and $\sigma_i$ of a latent normal distribution to the observation $x_i$. 
The challenge in IR is that algorithms face difficulty in learning to model rare values. Classical autoencoders also encounter this issue. In the section above, we proposed an initial level of adjustment, a balanced MSE, to enhance the VAE's learning process. However, the variance $\sigma_i$ assigned to rare values could still be significant, and the generated data may not faithfully represent the original data. Indeed, the VAE constructs a latent representation for each observation, i.e., calibrates the parameters of each latent variable independently: the latent representations are i.i.d.
It is important to emphasize that the data generation concerns rare values. The objective is indeed to construct a training sample consisting of "majority" values, observed, and rare values, observed or generated.

Here, we propose a second level of processing to improve the generation of rare data. 
First, we define the importance weights used for the drawing for rebalancing the learning process, and enabling better modeling of rare values, as follows:
\begin{align}\label{omega}
\omega_i := \frac{1}{\hat{f}_Y(y_i)^\alpha} 
\end{align}

We then suggest not generating conventionally with the VAE but rather through a smoothed bootstrap (see \citep{silverman1987bootstrap}, \citep{hall1989smoothing}, \citep{de1992smoothing}) on the $n\times q$ matrix of values $\mu$  representing the mean value of the new representation of $X$ in the latent space of dimensionality $q$: $\mathbf{\mu}:=\{\mu_{ij}, i=1,\cdots,n \, ; \, j=,\cdots,q\}$.


Smoothed bootstrap (SB) consists of drawing samples from kernel density estimators of the distribution. It can be decomposed into two steps: first, a seed is randomly drawn and second, a random noise from the kernel density estimator is added to obtain a new sample. Here, the first step is represented by the drawing weight $\omega_i$ and the second by a kernel generation on $\mu$. 
Convergence properties of smoothed bootstrap are studied in \citep{de2008multivariate} and \citep{falk1989weak}. They proved the consistency of the smoothed bootstrap with classical multivariate kernel estimator and more specifically the convergence in Mallows metric. 
We propose to use a mixture of multivariate kernel (e.g. Gaussian) to generate synthetic data: 
\begin{align*}
  g_{Z^*}(z^*| \mathbf{\mu}) = \displaystyle \sum_{i=1,\cdots,n} \omega_i K_i(z^*,\mathbf{\mu}),
\end{align*}
where $(K_i)_{i =1,\cdots,n}$ is a collection of kernel, $(\omega_i)_{i=1,\cdots,n}$ is a sequence of positive weights with
$\sum_{i=1,\cdots,n} \omega_i=1$. 
Here the index $^*$ stands for the synthetic data. 
The smoothed bootstrap has the advantage of not requiring any additional parameters. This method is based on the kernel density estimate technique, for which the estimation of the smoothing parameter $H$ is already optimally proposed. Indeed, \citep{scott2015multivariate} or \citep{silverman1986density} suggest a parametrization of the bandwidth matrix to obtain consistency. Moreover, It is important to note that the VAE provides a regular space, offering an ideal framework for the smoothed bootstrap. This may not be the case for the initial space, which can exhibit discontinuities and bounded distributions. Finally, as the Gausian Noise \citep{Lee2000GN} and ROSE \citep{Menardi2014ROSE} algoritms, we use a parameter tuning the level noise and manage the generation. This multiplicative parameter (<1) is applied directly to the smoothing matrix.

We could have applied this method to a classical autoencoder, but VAEs ensure the necessary continuity of the support for the application of smoothed bootstrap. Indeed, a kernel density estimator can be seen as a Gaussian mixture, just like the latent distribution of a VAE. However, a VAE generates from a Gaussian distribution specific to each observation, with possible biased parameters for rare values. Here, generation is based on the non-parametric estimation of the distribution of mean latent values $\mu_i$, and the variance used for generation is derived from the neighborhood and the distribution of means, rather than attempting to estimate a parameter specific to each observation ($\sigma_i$).
\\
More precisely, for the classical VAE each latent variable can be express as 
$z_i = \w \mu_i + \w \sigma_i \epsilon$, with  $\epsilon \sim \mathcal{N}(0,1) $. The generator associated to $z_i$ has the form 
$g_{z^*}(z^*|z_i) \sim \mathcal{N}(\w \mu_i,\w \sigma_i)$. 
The weakness of this generator in the presence of a rare $ z_i$ value is that it takes into account the associated variance $\si_i$, which is likely to be very large and  the associated  generated value of $z^*$ will often be  far from $\mu_i$.  
We are trying to remedy this problem with  DAVID, taking into account all the $\mu_i$  
and proposing a joint generative function as follows:
\begin{align*}
g_{z^*}(z^*|\mu)  = \sum_{i=1}^n \omega_i K_{H_n}(z^*-\mu_i), 
\end{align*}
where $K$ is a Gaussian kernel, $H_n := \eta \times Var(\mu)$ is defined by Silverman’s rule of thumb: $\eta = (\frac{4}{p+2})^{\frac{1}{d+4}} n^{\frac{-1}{d+4}}$ or Scott’s rule of thumb: $\eta = n^{\frac{-1}{d+4}}$ , $Var(\mu)$ being the variance matrix of $\mu$.
We notice that the generation of $z^*$ from a seed $\mu_i$ is performed based on the kernel centered on it, i.e., depending of its neighborhood and the characteristics of the distribution of $\mu$. The VAE proposes to generate from the inference made on the seed $z_i$, i.e., from a Gaussian distribution with parameters ($\mu_i$, $\sigma_i$). As mentioned earlier, these parameters of the latent Gaussians can be difficult to estimate for rare values, so it is wise to rely on the neighborhood of the latent representation of rare values to generate new values. Furthermore, the assumption of latent Gaussians can be too strong for tabular data (\citep{ma2020vaem}), so generating from a classical VAE may be ineffective in certain situations.

\subsection{Algorithm}

\begin{figure}[ht]
     \centering
         \includegraphics[width=\textwidth]{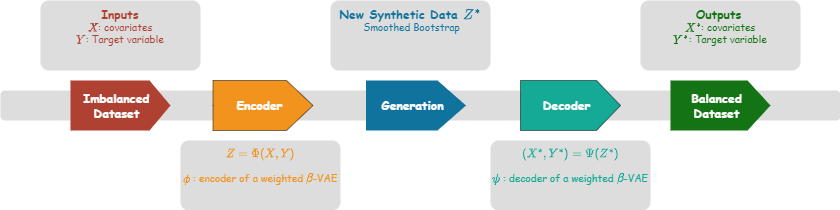}
         \caption{DAVID Algorithm: Data Generation}
         \label{DavidScheme}
\end{figure}

Our approach is based on two main steps: 
\begin{enumerate}
\itemsep0em 
    \item The first one is the $\beta-$VAE for regression training with the balanced loss function on the target variable.
    \item The second step involves generating synthetic data: 
    \begin{enumerate}
        \item[i)] by drawing a seed, the mean representation in the latent space of a rare initial observation $\mu_i$, obtained with the VAE's encoder;
        \item[ii)] by generating a new observation in the latent space $z^*$ using a smoothed bootstrap applied on $\mu$; 
        \item[iii)] by generating a new observation (\(x^*,y^*)\) from $z^*$ using the decoder.
    \end{enumerate}

\end{enumerate}
The data generation is illustrated in Figure \ref{DavidScheme}. 

\section{Numerical Illustration} \label{illustration}

\subsection{Dataset}
We first test our approach on a simple simulated dataset of size 3,000. To do this, we simulate 6 numerical features $X=X_j, j=1,\cdots,6$ and 1 target variable $Y$ as follows:
\bit
\small
\item Features $X$ : $X_1 \sim \mathcal{N}(0,2)$ ; $X_2 \sim \mathcal{N}(10,2)$ ; $X_3 \sim \mathcal{N}(0,5)$  ; $X_4 \sim \mathcal{N}(X_1^3,1)$ ; $X_5 \sim \mathcal{N}((X_2-10)^2,1)$ ; $X_6 \sim \mathcal{N}(X_3^2,2)$
\item Target variable $Y$ : $Y \sim \mathcal{N}(U^2,10)$ with $U:= 11 mm(X_4) + 9 mm(X_5) +14 mm(X_6) +10$ with $mm$ the Min-Max Scaler: $mm(X):=\frac{X-\min(X)}{\max(X)-\min(X)}$
\eit

As observed here, we simulated a dataset of 7 variables with nonlinear relationships. The dataset is defined based on 3 Gaussian latent variables to which a slight Gaussian noise is added. 

\subsection{Protocol}
We construct a training set and a test set through uniform random sampling, aiming to allocate 70\% of the data to the train. Next, we construct four models: a vanilla autoencoder, a 0-VAE (by canceling the penalty of the Kullback-Leibler divergence), a vanilla $\beta$-VAE, and a rebalanced $\beta$-VAE with weights on $Y$ given by \ref{Bal_loss}. It is important to note that a vanilla VAE, i.e., with a weight of 1 for the KL divergence, does not yield good results. Indeed, enforcing normality in the latent space penalizes the reconstruction of observations (\citep{ma2020vaem}).
Therefore, the parameter $\beta$ in the $\beta$-VAE is smaller than 1. 
The architectures of the models and the hyperparameters are detailed in the Appendix \ref{models_archi_illu}\footnote{Code and data are available at: \url{https://github.com/sstocksieker/DAVID/}}.

Once the models are constructed, we generate several synthetic samples from the same seed drawing with the weights defined previously \ref{omega}. To measure the impacts of our proposals, we propose a step-by-step study in the evaluation of our results.
We then obtain several sets of training data:
\bit 
\item Initial training set: \textit{Baseline}
\item Oversampling, applied on the training set, with the drawing weights \ref{omega}: \textit{OS}
\item Classical Smoothed bootstrap, applied on the training set i.e. in the data space: \textit{CSB}
\item Natural generation with the $0-VAE$: \textit{0VAE}
\item Natural generation with the vanilla $\beta-VAE$: \textit{BVAE}
\item Smoothed bootstrap from the $\mu$ distribution in latent space $Z$ obtained with the vanilla $\beta-VAE$: \textit{kBVAE}
\item Natural generation with the rebalanced $\beta-VAE$ i.e. loss function defined as \ref{Bal_loss}: \textit{BVAEw}
\item Smoothed Bootstrap from the $\mu$ distribution in latent space $Z$ obtained with the rebalanced $\beta-VAE$ i.e. loss function defined as \ref{Bal_loss}: \textit{kBVAEw}.  This approach will be referred to as the David generator, which combines all the steps of our algorithm. 
\item Smoothed bootstrap on the latent space obtained with PCA: \textit{kPCA}
\item Smoothed bootstrap on the latent space obtained with kernel-PCA (polynomial kernel): \textit{kKPCA}
\item We finally combine smoothed bootstrap with an autoencoder:  \textit{kAE}. This last method is to be avoided because the autoencoder does not guarantee the regularity of the latent space, which theoretically prevents the application of the SB.
\eit 

To compare our approach with the state of the art, we also simulate synthetic data using the following methods:
\bit
 \setlength{\itemsep}{0pt}
\item A Tabular Variational Autoencoder (\citep{ctgan}), trained with the same epochs than ours, from the python-package \textit{Synthetic Data Vault} (SDV) \citep{SDV}: \textit{TVAE}
\item A conditional Tabular Generative Adversarial Net (\citep{ctgan}), trained with the same epochs than ours, from the python-package \textit{Synthetic Data Vault} (SDV) \citep{SDV}: \textit{CTGAN} 
\item A Copula-Generative Adversarial Net, trained with the same epochs than ours, from the python-package \textit{Synthetic Data Vault} (SDV) \citep{SDV}: \textit{CopGAN} 
\item Oversampling, applied on the training set, with the UBL approach (\citep{branco2016ubl}) from the python-package \textit{imbalancedlearningregression} (\citep{wu2022imbalancedlearningregression}): \textit{ILRro}
\item The SMOTE for Regression (\citep{torgo2013smote}), applied on the training set, with the UBL approach (\citep{branco2016ubl}) from the python-package \textit{imbalancedlearningregression} (\citep{wu2022imbalancedlearningregression}): \textit{ILRsmote}
\item The Gaussian Noise for Regression (\citep{branco2017smogn}), applied on the training set, with the UBL approach (\citep{branco2016ubl}) from the python-package \textit{imbalancedlearningregression} (\citep{wu2022imbalancedlearningregression}): \textit{ILRgn}
\item The ADASYN method \citep{he2008adasyn}, with the UBL approach (\citep{branco2016ubl}) from the python-package \textit{imbalancedlearningregression} (\citep{wu2022imbalancedlearningregression}) was also applied but removed because the results were very poor and, more importantly, the computational time was too high.
\item We would like to note that we were unable to compare our results with the VIR \citep{wang2024variational} approach as their code is no longer accessible.
\eit

The next step involves predicting the target variable $Y$ of the test set using the different training datasets, including, as the baseline, the initial training dataset. To obtain robust results, we apply the comparison on 10 train-test samples (K-fold approach). It is important to note that the train sets are mixed, meaning they are constructed by blending original data (especially for frequent observations) and synthetic data (especially for rare values). In the same way, to avoid getting results dependent on some learning algorithms we use 10 models from the \textit{autoML of the H2O package} \citep{H2OAutoML20} among the following algorithms: Distributed Random Forest, Extremely Randomized Trees, Generalized Linear Model with regularization, Gradient Boosting Model, Extreme Gradient Boosting and a Fully-connected multi-layer artificial neural network. 

\subsection{Results}
We compare the prediction results with the following metrics:
\bit
 \setlength{\itemsep}{0pt}
\item The standard Mean Squared Error $MSE(Y,\w Y) := \frac{1}{n} \sum_{i=1}^n (y_i - \w y_i)^2$
\item Our weighted MSE $wMSE(Y,\w Y) := \frac{1}{n} \sum_{i=1}^n \omega_i (y_i - \w y_i)^2$ with $\omega_i$ defined as in \ref{omega}
\item The standard Mean Absolute Error $MAE(Y,\w Y) := \frac{1}{n} \sum_{i=1}^n |y_i - \w y_i|$
\item The Mean Absolute Percentage Error $MAPE(Y,\w Y) := \frac{1}{n} \sum_{i=1}^n \frac{|y_i - \w y_i|}{\max(\epsilon, |y_i|)}$
\eit


\begin{table}[ht]
\centering 
\resizebox{0.8\columnwidth}{!}{%
\begin{tabular}{|c||c|c|c|c|}
 \hline
 Metric & MSE & wMSE & MAE & MAPE   \\
 \hline
 Train	&mean (std)	&mean (std)	&mean (std)	&mean (std)\\
 \hline
Baseline        &222	(29)	&0,11	(0,01)	&7,78	(0,21)	&0,06	(0,00) \\
OS              &245	(36)	&0,12	(0,02)	&7,95	(0,18)	&0,06	(0,00) \\
CSB             &247	(24)	&0,12	(0,01)	&8,11	(0,16)	&0,06	(0,00) \\
kAE             &251	(31)	&0,12	(0,01)	&8,10	(0,22)	&0,06	(0,00) \\
OVAE            &220	(24)	&0,10	(0,01)	&7,77	(0,19)	&0,06	(0,00) \\
BVAE            &218	(23)	&0,10	(0,01)	&7,79	(0,18)	&0,06	(0,00) \\
kBVAE (ours)	&218	(23)	&0,10	(0,01)	&7,79	(0,18)	&0,06	(0,00) \\
BVAEw (ours)	&214	(25)	&0,10	(0,01)	&7,81	(0,14)	&0,06	(0,00) \\
\textbf{kBVAEw (ours)}	&\textbf{202	(22)}	&\textbf{0,10	(0,01)}	&\textbf{7,70	(0,18)}	&\textbf{0,05	(0,00)} \\
kPCA (ours)     &247	(28)	&0,12	(0,01)	&8,10	(0,19)	&0,06	(0,00) \\
kKPCA (ours)	&494	(258)	&0,24	(0,12)	&9,42	(2,22)	&0,07	(0,02) \\
TVAE            &214	(23)	&0,10	(0,01)	&7,72	(0,15)	&0,06	(0,00) \\
CTGAN           &297	(47)	&0,14	(0,02)	&8,24	(0,31)	&0,06	(0,00) \\
CopGAN          &344	(82)	&0,16	(0,04)	&8,83	(1,39)	&0,07	(0,01) \\
ILRro           &269	(39)	&0,13	(0,02)	&8,13	(0,27)	&0,06	(0,00) \\
ILRsmote        &245	(32)	&0,12	(0,02)	&7,97	(0,21)	&0,06	(0,00) \\
ILRgn           &238	(24)	&0,11	(0,01)	&8,20	(0,23)	&0,06	(0,00) \\
 \hline
\end{tabular}
}%
\caption{Illustration results: metrics for test prediction}
\label{table_illu}
\end{table}

The results given in Table \ref{table_illu}  show that:
\begin{itemize}
    \item The results of the \textit{BVAE} are slightly better than those of the \textit{0VAE}, i.e., this shows the importance of having a Kullback-Leibler penalty, which notably helps obtain a regular latent space.
    \item The results of the \textit{kAE} do not appear to be better than those of the \textit{kTrain}: this shows that embedding the data into a latent space to apply the smoothed bootstrap is not sufficient: the latent space must have regularity properties
    \item The results of \textit{kBVAE} and \textit{BVAE} are similar: generation with a smoothed bootstrap gives the same results as the natural generation of the BVAE
    \item The results of the \textit{BVAEw} are better than those obtained from the \textit{BVAE}: this demonstrates the relevance of the proposed loss function.
    \item The results of DAVID, that is, the \textit{kBVAEw}, are better than those of the \textit{BVAEw}: generating data with a smoothed bootstrap is more suitable than natural generation of the VAE.
    \item The results of \textit{kPCA} and \textit{kKPCA} are quite poor: applying a smoothed bootstrap to the latent space of factorial approaches is not relevant.
    \item Synthetic data generators that are not specific to IR (\textit{TVAE}, \textit{CTGAN}, \textit{CopGAN}) are not suitable: the results obtained are worse than those obtained with the initial sample.
    \item Finally, DAVID outperforms the state-of-the-art methods (\textit{ILRro},\textit{ILRsmote},\textit{ILRgn}).
\end{itemize}

A study on performance ranking shown that DAVID is often superior and that the results are quite robust.


\section{Experiments} \label{Experiments}

To test our approach on real datasets, we compare the results on benchmark datasets for IR taken from \citep{branco2019pre}\footnote{The dedicated repository "Data Sets for Imbalanced Regression Learning" is available at this address: \url{https://paobranco.github.io/DataSets-IR/} }.

\begin{table}[ht]
\centering
\resizebox{0.8\columnwidth}{!}{%
\begin{tabular}{|c||c|c|c|c|}
 \hline
 Dataset &bank8FM	&abalone	&boston	&NO2   \\
 \hline
 Train	&mean (std)	&mean (std)	&mean (std)	&mean (std)\\
 \hline
Baseline &1,87 (0,19) &6,84 (0,29) &44,44 (9,57) &0,48 (0,04) \\
OS &2,61 (0,4) &6,19 (0,17) &41,01 (11,07) &0,47 (0,06) \\
kTrain &3,3 (0,52) &5,81 (0,18) &37,49 (12,47) &0,46 (0,07) \\
kAE &1,77 (0,16) &5,79 (0,19) &37,53 (11,62) &0,44 (0,05) \\
OVAE &1,78 (0,15) &5,78 (0,25) &40 (9,9) &0,42 (0,04) \\
BVAE &1,76 (0,16) &5,74 (0,19) &38,99 (11,12) &0,42 (0,04) \\
kBVAE (ours) &1,74 (0,17) &5,6 (0,19) &37,37 (12,03) &0,43 (0,04) \\
BVAEw (ours) &1,71 (0,16) &5,67 (0,22) &35,67 (11,82) &0,42 (0,04) \\
\textbf{kBVAEw (ours)} & \textbf{1,68 (0,13)} & \textbf{5,52 (0,22)} & \textbf{34,86 (12,12)} & \textbf{0,4 (0,04)} \\
kPCA (ours) &3,35 (0,52) &5,83 (0,21) &39,22 (11,64) &0,46 (0,07) \\
kKPCA (ours) &2,57 (0,51) &5,81 (0,18) &38,99 (11,06) &0,44 (0,05) \\
TVAE &2,56 (0,65) &7,17 (0,33) &45,2 (12,59) &0,5 (0,06) \\
CTGAN &8,96 (4,55) &7,43 (0,32) &51,77 (15,11) &0,54 (0,05) \\
CopGAN &8 (3,08) &7,5 (0,3) &55,79 (12,37) &0,57 (0,08) \\
ILRro &3,37 (0,36) &6,42 (0,2) &46,11 (10,85) &0,48 (0,04) \\
ILRsmote &2,6 (0,33) &5,78 (0,2) &44,3 (9,19) &0,47 (0,04) \\
ILRgn &2,51 (0,38) &6,41 (0,18) &42,78 (10,73) &0,46 (0,04) \\
 \hline
\end{tabular}
}%
\caption{Experiments results: wMSE for test prediction}
\label{table_res_Xps}
\end{table}
The results, in Table \ref{table_res_Xps}, confirm those of the illustration, DAVID ("kBVAEw") gives better results than the initial training sample and the state-of-the-art approaches. The results show that it is preferable to generate synthetic data from the latent space rather than from the initial space ("kTrain"). Finally, these experiments show that using standard synthetic data generators ("TVAE", "CTGAN", "copGAN") is not recommended in the case of imbalanced regression. 

\section{Discussion and Perspectives} \label{Discussion}

This paper proposes an enhancement for learning in imbalanced regression (IR), which remains a relatively unexplored problem compared to classification, especially for structured data. We suggest embedding observations into a latent space to enable more relevant data generation than in the initial space. We empirically demonstrated that intuitive techniques like PCA and kernel-PCA do not offer a satisfactory framework for data augmentation. On the other hand, based on neural networks, deep learning approaches used effectively on images (deep imbalanced regression) are limited for tabular data. However, VAEs offer a variational inference capable of capturing nonlinear correlations, which is crucial in data generation. 

We propose here leveraging the power of VAE inference while reconsidering how to generate data. Morevoer, VAEs provide a suitable framework for the application of KDEs because they offer a regular latent space, which may not be the case in the original space or with a vanilla autoencoder. Our results empirically show that modifying only the loss function of a VAE for data generation in IR (in-processing) is satisfactory. But, we can improve this data generation by replacing the native data generation of the VAE with a smoothed bootstrap. The underlying idea is to use the neighborhood of observations in the latent space to more effectively generate rare values, which are inherently rare i.e. difficult to model. It is interesting to note that generation from VAEs is semi-parametric (as it stems from latent Gaussians). The DAVID algorithm offers non-parametric generation based on a smoothed bootstrap, i.e., a KDE-like approach. Our simple and effective DAVID algorithm offers better results than conventional approaches in IR and than the traditional VAE on multiple datasets with various learning algorithms. Our study also demonstrates that using "vanilla" synthetic data generators (\textit{TVAE} or \textit{CTGAN}) not specifically dedicated to the problem of IR are not effective.

The relevance of our approach can be explained by its ability to consider nonlinear correlations between variables. This methodology is effective when the latent representation space accurately reflects the observations, meaning the VAE (Variational Autoencoder) must effectively reproduce the data and provide the continuity properties needed for applying smoothed bootstrap. Thus, the method would not work if the VAE is not functioning properly. Changing the representation space of the data to generate synthetic data is promising. It allows for the consideration of nonlinear correlations as well as mixed data, which remains a significant challenge  \citep{ma2020vaem}.

Algorithm DAVID could potentially be applied in the context of imbalanced classification, but this issue has already been extensively addressed with a plethora of solutions. Applying this approach to mixed tabular in IR data would be intriguing, as it remains a relatively unexplored and challenging application type, especially due to correlations associated with qualitative data. Finally, it would be interesting to test this new approach on images within the context of deep imbalanced regression.



%
%
%
%

\section*{Acknowledgment}

The authors thank the reviewers for their helpful comments which helped to improve the manuscript. S. Stocksieker would like to acknowledge the support of the Research Chair DIALog under the aegis of the Risk Foundation, a joint initiative by \textit{CNP Assurance}. D. Pommeret would like to acknowledge the support received from the Research Chair ACTIONS under the aegis of the Risk Foundation, an initiative by BNP Paribas Cardif and the Institute of Actuaries of France.

\bibliography{references}


\appendix

\section{Model Architecure and Parameters} \label{models_archi_illu}

For the illustration and experiments, we have chosen the following parameters:
\begin{itemize}
    \item train-test size: 60-40\% 
    \item The $\alpha$ parameter for weighting: $1$. This parameterization allows convergence to a continuous uniform target distribution, i.e. all observations are given equal weight.
    \item The parameter tuning the level noise for perturbation approaches: $0.1$. This parameter ensures that we do not stray too far from observing the seed in latent space. It is a classical default value in equivalent approaches \citep{branco2016ubl}. 
    \item Epochs: $2000$, batch size: $128$, learning rate: $10e^{-3}$
    \item The weights of $X$ reconstruction in loss function: $\beta_x=1$. This parameter is always set to 1. Setting the weight of the Y is sufficient.  
    \item The weights of $Y$ reconstruction in loss function: $\beta_y=10$. The weight of the $Y$ reconstruction is approximately equal to the number of covariates. 
    \item The weights of Kullback-Leibler term in loss function: $\beta_KL=1e^{-6}$. It is important to highlight that a standard VAE, which assigns a weight of 1 to the KL divergence, doesn't produce satisfactory results. This is because enforcing normality in the latent space hinders the reconstruction of observations \citep{ma2020vaem}. Consequently, the $\beta$ parameter in the $\beta$-VAE is set to less than 1.
    \item The Kernel Density Estimate for Smoothed Boostrap and the weighting in loss function and drawing is obtained with a Gaussian Kernel with Silverman'rule (using the python-package \textit{Scipy}, as described in \ref{smb_generation})
\end{itemize}
The architecture of our $\beta-$VAE is as follows:
\begin{itemize}
    \item an encoder $\phi$ consisting of 5 hidden layers of dimensions \\
    $(p+1,2p+1,p-q,p-2q,p-3q/p-3q)$
    \item a decoder $\psi$ consisting of 5 hidden layers of dimensions \\
    $(p-3q,p-2q,p-q,2p+1,p/1)$
    \item the parameter of HL dimension reduction $q=int(p/10)+1$
    \item activation function are all $Tanh()$
    \item a standard reparametrization based on a random standard Gaussian 
    
\end{itemize}

\section{Dataset Details}

To test our approach on real datasets, we compare the results on benchmark datasets for IR taken from \citep{branco2019pre}\footnote{The dedicated repository "Data Sets for Imbalanced Regression Learning" is available at this address: \url{https://paobranco.github.io/DataSets-IR/}}. The datasets are composed as follows:
\bit
\item The \textit{bank8fm} dataset: 4498 observations and 9 variables (7 float and 1 integer). 
\item The \textit{abalone} dataset: 4177 observations and 9 variables (8 float and 1 integer). 
\item The \textit{boston} dataset: 506 observations and 14 variables (11 float and 3 integer).
\item The \textit{NO2} dataset: 4498 observations and 9 variables (8 float and 1 integer). 
\eit

\end{document}